\documentclass{preprint} 

\title[M{\"o}bius Learning]{M{\"o}bius Learning: Cyclic Depth Folding in Transformers\\[0.35em]
{\large\textcolor{black!38}{Version 1 (Preliminary Technical Report)}}}

\paperauthor{%
\Name{Tongtian Zhu}\\
\addr Zhejiang University \\
\texttt{raiden@zju.edu.cn}}

\usepackage[utf8]{inputenc}
\usepackage[T1]{fontenc}
\usepackage{booktabs}
\usepackage{amsfonts}
\usepackage{nicefrac}
\usepackage{microtype}
\usepackage{colortbl}
\usepackage{dsfont}
\usepackage{mathtools}
\usepackage[shortlabels]{enumitem}
\usepackage{mdframed}
\usepackage[normalem]{ulem}
\usepackage[most]{tcolorbox}
\usepackage{tikz}
\usepackage{pgfplots}
\usepackage{svg}
\usepackage{caption}
\usepackage{subcaption}
\usepackage{placeins}
\usepackage{multicol,multirow,diagbox}
\usepackage{fontawesome5}
\usepackage{footmisc}
\usepackage[capitalize,nameinlink]{cleveref}
\usepackage[textsize=tiny]{todonotes}
\usepackage{pifont}

\setlist[itemize]{topsep=0em, itemsep=0em, partopsep=0em, parsep=0.5em}
\allowdisplaybreaks
\usetikzlibrary{backgrounds}
\usetikzlibrary{arrows,shapes}
\usetikzlibrary{tikzmark}
\usetikzlibrary{calc}
\pgfplotsset{compat=1.18}

\colorlet{LightBlue}{blue!39!white}
\colorlet{DarkBlue}{blue!70!black}
\colorlet{VeryLightBlue}{blue!30!white}
\colorlet{LightRed}{red!35!white}

\definecolor{darkgrey}{rgb}{0.53,0.53,0.53}
\definecolor{middlegrey}{rgb}{0.75,0.75,0.75}
\definecolor{mygrey}{rgb}{0.9,0.9,0.9}
\definecolor{mydarkblue}{rgb}{0,0.08,0.45}
\definecolor{darkdarkblue}{rgb}{0.0,0.0,0.3}
\definecolor{darkblue}{rgb}{0.0,0.0,0.7}
\definecolor{darkred}{rgb}{0.4,0,0.3}
\definecolor{lightblue}{HTML}{F9FEFE}
\definecolor{verylightpurple}{HTML}{FBFAFF}
\definecolor{lightred}{HTML}{FFFAFA}
\definecolor{fancyblue}{HTML}{4771E3}
\definecolor{grey}{rgb}{0.95,0.95,0.95}
\definecolor{myred}{HTML}{7A1410}
\definecolor{lightpurple}{HTML}{ECE5F3}
\definecolor{myorange}{HTML}{FFDD81}
\definecolor{mypink}{HTML}{ffaec9}
\definecolor{boxcyan}{RGB}{225, 245, 250}
\definecolor{textcyan}{RGB}{0,78,181}
\definecolor{boxpink}{RGB}{250, 225, 245}
\definecolor{textpink}{RGB}{120, 0, 100}
\definecolor{PlotBlue}{HTML}{4A90E2}
\definecolor{PlotOrange}{HTML}{F5A542}
\definecolor{PlotBackground}{HTML}{F5F5F9}
\definecolor{PlotGrid}{HTML}{FFFFFF}
\definecolor{PlotFrame}{HTML}{C9CBD3}
\definecolor{TableHeaderPurple}{HTML}{ECE9F1}
\definecolor{TableStructuralPurple}{HTML}{F8F7FA}
\definecolor{TableRuleGray}{HTML}{626066}

\newtcolorbox{questionbox}{
  enhanced,
  notitle,
  rounded corners,
  colframe=middlegrey,
  colback=lightblue,
  boxrule=2pt,
  boxsep=0pt,
  left=0.15cm,
  right=0.17cm,
  toprule=2pt,
  before skip=0.75em,
  after skip=0.75em
}

\hypersetup{
    colorlinks=true,
    linkcolor=darkred,
    citecolor=darkblue,
    filecolor=darkblue,
    urlcolor=darkblue
}

\crefname{definition}{Definition}{Definitions}
\crefname{assumption}{Assumption}{Assumptions}
\crefname{theorem}{Theorem}{Theorems}
\crefname{remark}{Remark}{Remarks}
\crefname{lemma}{Lemma}{Lemmas}
\crefname{corollary}{Corollary}{Corollaries}
\crefname{proposition}{Proposition}{Propositions}
\crefname{claim}{Claim}{Claims}
\crefname{exercise}{Exercise}{Exercises}
\crefname{section}{Section}{Sections}
\crefname{subsection}{Subsection}{Subsections}
\crefname{example}{Example}{Examples}
\crefname{table}{Table}{Tables}
\crefname{problem}{Problem}{Problems}
\crefname{algorithm}{Algorithm}{Algorithms}
\crefname{figure}{Figure}{Figures}
\crefname{property}{Property}{Properties}
\crefformat{equation}{#2Equation~(#1)#3}
\crefformat{inequality}{#2Inequality~{}(#1){}#3}
\crefrangeformat{inequality}{#3Inequality~(#1)#4--#5(#2)#6}
\Crefformat{section}{#2Section~#1#3}
\Crefformat{page}{#2Page~#1#3}

\newcommand{\acref}[1]{\hyperref[#1]{Appendix~\ref*{#1}}}

\newcommand{\method}{M{\"o}bius Learning}

\begin{document}
\maketitle

\begin{center}
  \makebox[\linewidth][l]{%
  \begin{minipage}[t]{0.50\linewidth}
  \vspace{0pt}
  \raggedright
  \includegraphics[width=1.03\linewidth]{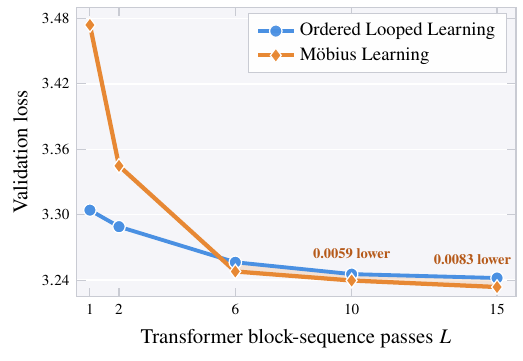}
  \end{minipage}%
  \hspace{0.035\linewidth}%
  \begin{minipage}[t]{0.43\linewidth}
  \vspace{1.5pt}
  \centering
  \includegraphics[width=\linewidth]{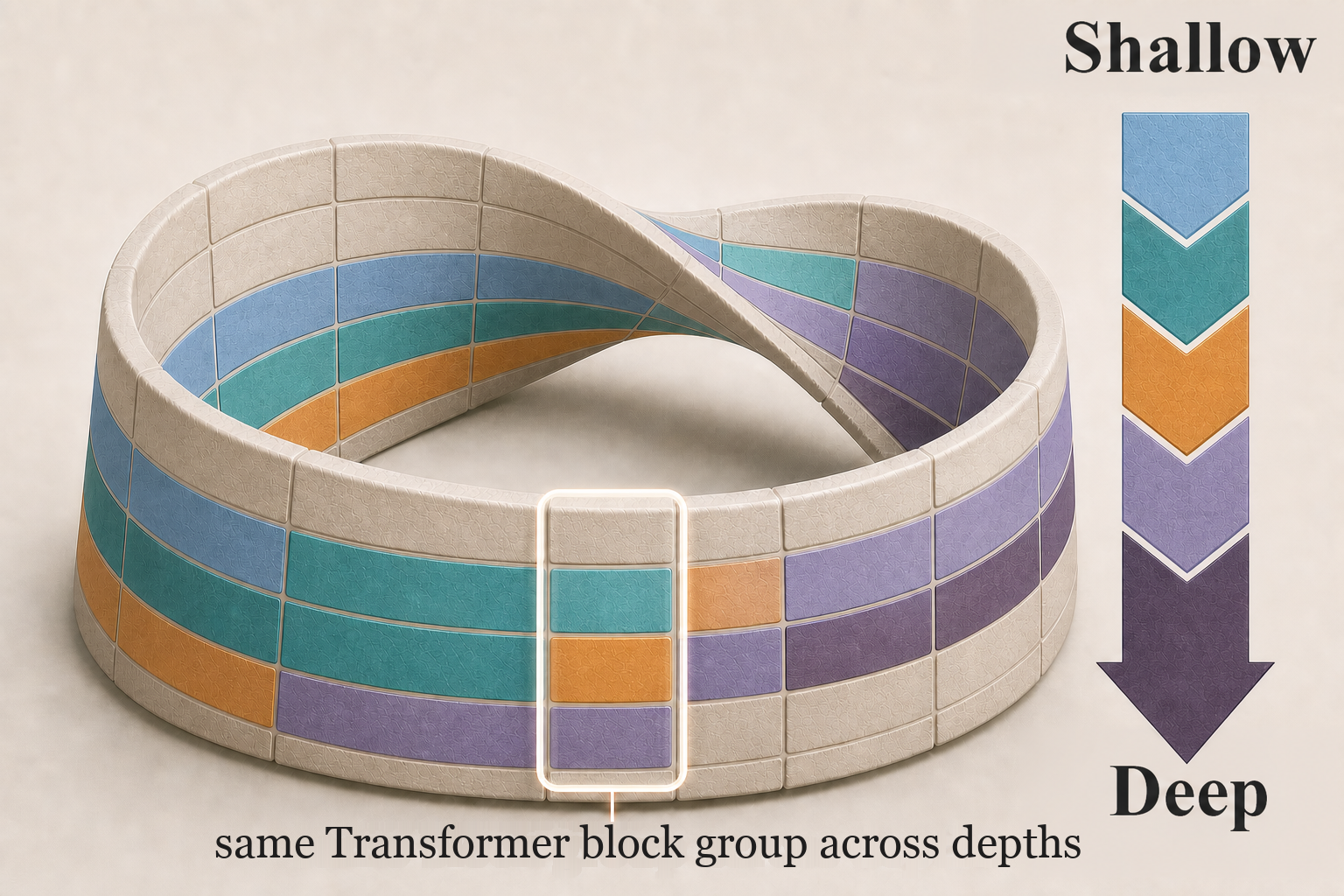}
  \end{minipage}%
  }
  {\captionsetup{type=figure}
  \caption{\textbf{Left:} FineWeb validation loss after 2.5B training tokens for four-worker modded GPT-2 small (124M) runs trained with Muon. Ordered-loop training and \method{} are compared at the same loop depth $L$, defined as the number of complete Transformer block-sequence passes. With $P=4$ block groups, each setting therefore performs $K=4L$ block-group applications. At loop depths $L=6,10,15$, \method{} achieves lower validation loss than ordered-loop training. \textbf{Right:} conceptual illustration of cyclic depth folding. The colored bands denote depth roles, and the highlighted block group can serve at shallow and deep positions across cyclic orders.}
  \label{fig:gb512-validation}
  }
\end{center}

\begin{abstract}
Transformer-based language models organize computation along an ordered depth axis, where shallow and deep blocks often develop distinct representational roles. We challenge the conventional view that these roles must remain tied to a block's position in the ordered sequence. We introduce \method{}, a training architecture based on cyclic depth folding, in which different data streams follow cyclically shifted block orders. The same block group is therefore applied early in the block sequence for some data streams and late for others, so it is optimized in both shallow and deep roles, a phenomenon we call depth-role superposition. Surprisingly, in four-worker experiments with a modded GPT-2 small (124M) model trained on 2.5B FineWeb tokens using Muon, \method{} achieves lower validation loss than a fixed-order looped Transformer at larger numbers of Transformer block-sequence passes. This counterintuitive result shows that a block group need not remain confined to one fixed shallow or deep role within the block sequence and opens a new design space based on cyclic depth folding. Crucially, this structure makes \method{} particularly well suited to memory-constrained distributed training: raw training data remain local, while each worker stores one block group rather than the complete Transformer block stack.
\end{abstract}

\section{Introduction}
\label{sec:introduction}

In Transformer-based language models, a block's position determines the representational context in which it operates: shallow blocks transform states produced early in the sequence, whereas deep blocks transform states already shaped by many preceding transformations~\citep{vaswani2023attentionneed,radford2019language}. Layer-wise studies report corresponding functional differences: linguistic features, attention behavior, and feed-forward memories vary across depth, suggesting that shallow and deep blocks contribute differently as representations evolve through the network~\citep{tenney2019bertrediscoversclassicalnlp,voita2019bottomupevolutionrepresentationstransformer,clark2019doesbertlookat,geva2021transformerfeedforwardlayerskeyvalue}. This depth--function association makes training the same block group at shallow and deep positions in the block sequence a nontrivial optimization problem.

Existing flexible-depth Transformers largely follow two strategies while retaining a common block order. Recurrent and cross-layer-sharing models reuse parameters across computational depth~\citep{dehghani2019universaltransformers,lan2020albertlitebertselfsupervised,reid2021subformerexploringweightsharing}, whereas depth-reduction and token-adaptive methods vary the amount of computation at the model or token level~\citep{fan2019reducingtransformerdepthdemand,raposo2024mixtureofdepthsdynamicallyallocatingcompute}. Among parameter-reuse approaches, fixed-order sharing schemes and recent looped language models repeatedly apply shared computation along recurrent depth, yet retain a model-wide prescribed block order~\citep{takase2023lessonsparametersharing,saunshi2025reasoninglatentthoughtspower,geiping2025scalingtesttimecomputelatent,zhu2025scalinglatentreasoninglooped,prairie2026parcaescalinglawsstable}. We use \emph{loop depth} to denote the number of complete Transformer block-sequence passes. Repeated passes therefore increase computational depth without changing a block group's relative position within the shared order.

\begin{questionbox}
\textbf{Question:} \textit{\fontsize{9.5pt}{12pt}\selectfont Can the same Transformer block group be trained to serve both shallow and deep roles?}
\end{questionbox}

\method{} opens this missing degree of freedom through cyclic depth folding. The Transformer is partitioned into block groups, and each worker introduces its local data at its own block group, fixing the cyclic starting point for that data stream. Different streams therefore traverse the same groups in cyclically shifted orders. The same block group is applied early in the sequence for some streams and late for others, so its parameters are optimized in both shallow and deep roles, a phenomenon we call depth-role superposition. We define a block group's source-relative depth for a given data stream as its position within a pass, measured from the stream's starting point, and formalize this construction in \cref{sec:methodology}.

\begin{figure}[!t]
  \centering
\includegraphics[width=\linewidth]{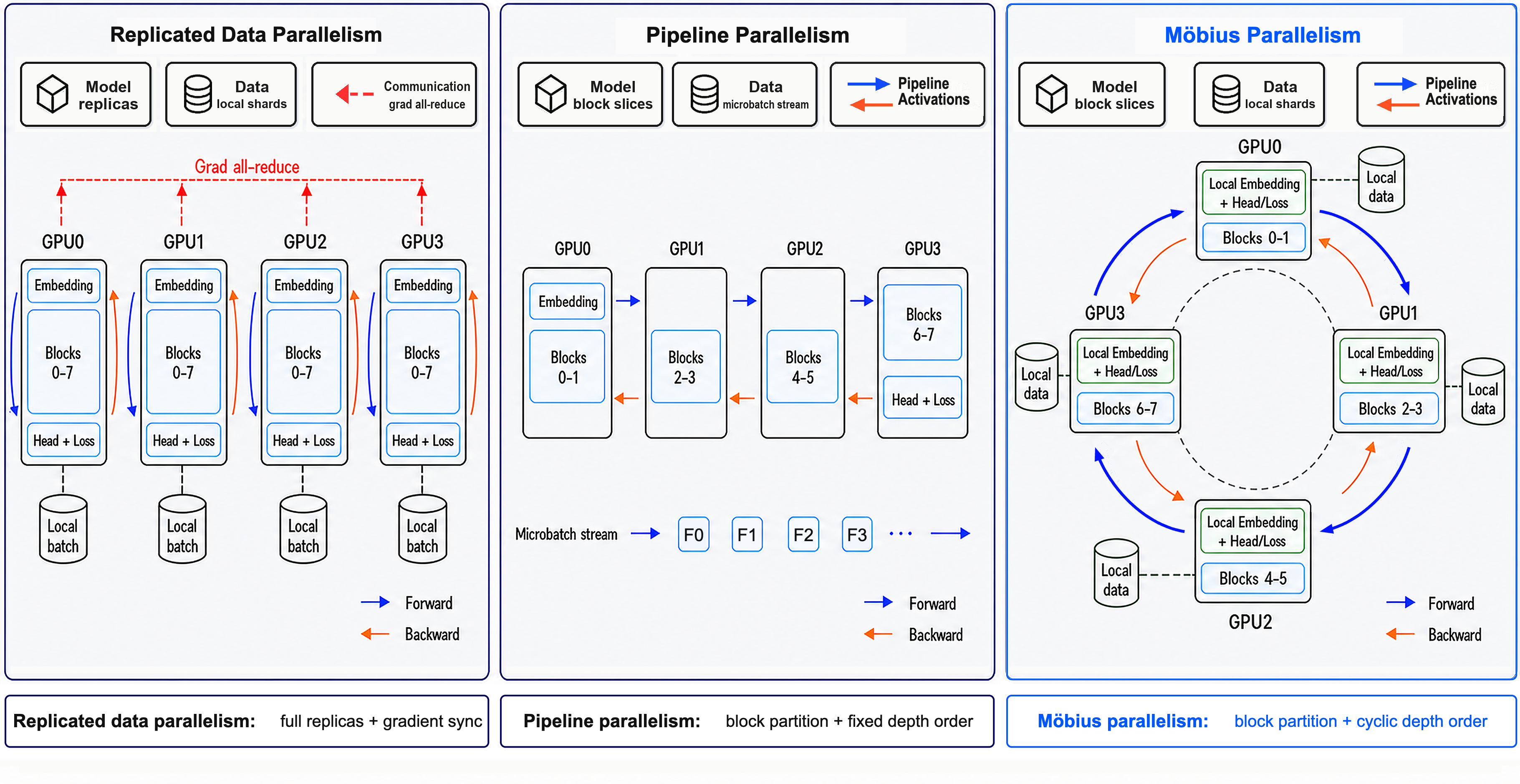}
  \caption{Conceptual comparison of three forms of parallelism. Replicated data parallelism retains a full ordered model at each worker, while pipeline parallelism partitions the model under a fixed stage order. In M{\"o}bius parallelism, each worker stores one block group and begins the cycle there, allowing the same group to serve at different depths for data from different workers.}
  \label{fig:distributed-comparison}
\end{figure}

We use \emph{M{\"o}bius parallelism} to denote the distributed execution strategy induced by \method{}. \Cref{fig:distributed-comparison} compares this strategy with replicated data parallelism and pipeline parallelism. Replicated data parallelism assigns each worker a full ordered model, whereas pipeline parallelism partitions the model across a fixed stage order~\citep{huang2019gpipeefficienttraininggiant,harlap2018pipedreamfastefficientpipeline}. In M{\"o}bius parallelism, each worker stores one block group and local data, while hidden states circulate through worker-indexed cyclic orders. Workers begin at different block groups, allowing the same group to serve at different depths.

These worker-indexed cyclic orders also determine where each block group operates within the block sequence for each data stream. The central optimization question is whether learning signals from the resulting shallow and deep roles interfere. We test this question in four-worker experiments using a modded GPT-2 small (124M) model~\citep{radford2019language} trained on 2.5B FineWeb tokens~\citep{penedo2024finewebdatasetsdecantingweb}. We compare \method{} with a fixed-order looped Transformer at the same loop depth, referring to this baseline as ordered-loop training. Surprisingly, \cref{fig:gb512-validation} shows that \method{} achieves lower validation loss than ordered-loop training at loop depths of 6, 10, and 15 passes.

LayerShuffle fine-tunes fixed-order pretrained Vision Transformers for robustness to shuffled block execution~\citep{freiberger2024layershuffleenhancingrobustnessvision}. \method{} instead trains cyclic orders throughout language-model pretraining, jointly optimizing the language-modeling objective as every block group serves shallow and deep roles; \acref{app:layer-order-related-work} compares the methods across loop depths.

\Needspace{8\baselineskip}
Cyclic depth folding changes both how Transformer depth is represented and how Transformer blocks are distributed. Shallow and deep roles coexist within the same block parameters, removing the usual one-to-one association between parameter groups and depth roles. Compared with replicated data parallelism, each worker stores one block group rather than the complete Transformer block stack, reducing per-worker memory for Transformer block parameters and enabling training when the full stack does not fit on one device.

\begin{samepage}
Our contributions are:
\begin{itemize}
    \item \textbf{Training architecture.} We introduce \method{}, a new training architecture that replaces the shared Transformer block order with worker-indexed cyclic orders. These orders make block depth depend on each data stream's starting point, while assigning one block group to each worker reduces per-worker memory for Transformer block parameters relative to replicated data parallelism.
    \item \textbf{Depth-role superposition.} We identify depth-role superposition: the same block parameters can be trained at shallow and deep source-relative positions, allowing them to serve both depth roles. A fixed source-relative depth for each block group is therefore not necessary for effective Transformer training.
    \item \textbf{Empirical result.} We compare \method{} and ordered-loop training under the same 2.5B-token budget and loop depth. At larger tested loop depths, training the same parameters across shallow and deep roles does not compromise validation performance: \method{} even achieves lower validation loss than ordered-loop training at 6, 10, and 15 passes.
\end{itemize}
\end{samepage}

\section{Methodology}
\label{sec:methodology}

\method{} assigns one Transformer block group to each worker. The worker at which a batch originates determines that batch's cyclic starting point. The input embedding associated with that path is applied once, its hidden states circulate through the assigned block groups in worker-indexed cyclic order, and the corresponding prediction map is applied once at the end. Different originating workers therefore induce cyclic shifts of the same block groups.

\subsection{Ordered and Looped Transformer Maps}
\label{subsec:ordered-transformer-maps}

Let $\mathcal{X}$ and $\mathcal{Z}$ denote token-sequence inputs and targets, $\mathcal{H}$ the hidden-state space, and $\mathcal{G}$ the vocabulary-logit space. Let $P\geq 1$ block groups be indexed by $\mathbb{Z}_P=\{0,\ldots,P-1\}$, and let $\langle a\rangle_P$ denote the unique element of $\mathbb{Z}_P$ congruent to $a$ modulo $P$; for example, $\langle-1\rangle_P=P-1$. An ordered Transformer consists of an input embedding $E:\mathcal{X}\to\mathcal{H}$, compatible block-group maps $F_i:\mathcal{H}\to\mathcal{H}$, and a prediction map $R:\mathcal{H}\to\mathcal{G}$. We use $(A\circ B)(x)=A(B(x))$ for function composition. One ordered Transformer block-sequence pass is
\begin{equation}
    \Psi
    =
    F_{P-1}\circ F_{P-2}\circ\cdots\circ F_0 .
    \label{eq:ordered-map}
\end{equation}
Thus, $F_0$ is applied first and $F_{P-1}$ last.

For an integer $L\geq 1$, ordered-loop training repeats this complete pass:
\begin{equation}
    \Psi^L
    =
    \underbrace{\Psi\circ\cdots\circ\Psi}_{L\ \mathrm{times}},
    \qquad
    R\circ\Psi^L\circ E:\mathcal{X}\to\mathcal{G}.
    \label{eq:ordered-loop-map}
\end{equation}
We call $L$ the \emph{loop depth}: the number of complete Transformer block-sequence passes. Across these passes, one block group is applied at each of $K=LP$ successive positions. The input embedding and prediction map are each applied once and are not counted in $L$ or $K$. Numbering the block-group positions from zero, $F_i$ occurs at $i+\ell P$ for $\ell=0,\ldots,L-1$, independently of the worker from which a batch originates. The case $L=1$ recovers the standard ordered stack.

\subsection{Cyclic Depth Folding}
\label{subsec:mobius-composition}

\method{} uses $P$ workers indexed by $\mathbb{Z}_P$, with worker $i$ owning block group $F_i$. For a batch originating at worker $s$, the index $s$ is the \emph{originating-worker index}. We denote by $E_s:\mathcal{X}\to\mathcal{H}$ and $R_s:\mathcal{H}\to\mathcal{G}$ the input embedding and prediction map invoked by the corresponding path.

For a batch input $x_s\in\mathcal{X}$ originating at worker $s$ and any block-group application count $K\geq 1$, initialize $h_{s,0}=E_s(x_s)$ and define
\begin{equation}
    h_{s,t+1}
    =
    F_{\langle s+t\rangle_P}(h_{s,t}),
    \qquad
    t=0,\ldots,K-1 .
    \label{eq:mobius-state-recursion}
\end{equation}
The recurrence applies the block groups in worker-indexed cyclic order. Equivalently,
\begin{equation}
    \Phi_{s,K}
    =
    F_{\langle s+K-1\rangle_P}
    \circ\cdots\circ
    F_{\langle s+1\rangle_P}
    \circ F_s ,
    \label{eq:mobius-map}
\end{equation}
and the resulting vocabulary logits are $R_s(\Phi_{s,K}(E_s(x_s)))$. Hence, $E_s$ and $R_s$ are each applied exactly once, regardless of $K$.

Although \cref{eq:mobius-map} is well defined for arbitrary $K$, the subsequent analysis and all reported experiments use complete passes, for which $K=LP$. Let $\Psi_s=\Phi_{s,P}$ denote one complete cyclic pass beginning at worker $s$. Periodicity of the worker-indexed cyclic order gives $\Phi_{s,LP}=\Psi_s^L$. In particular, $\Psi_0=\Psi$ and $\Phi_{0,LP}=\Psi^L$. For $s\neq 0$, $\Psi_s$ is a cyclic shift of the ordered pass and generally differs from $\Psi$, since the block-group maps need not commute.

\begin{figure}[!ht]
    \centering
    \includegraphics[width=0.96\linewidth]{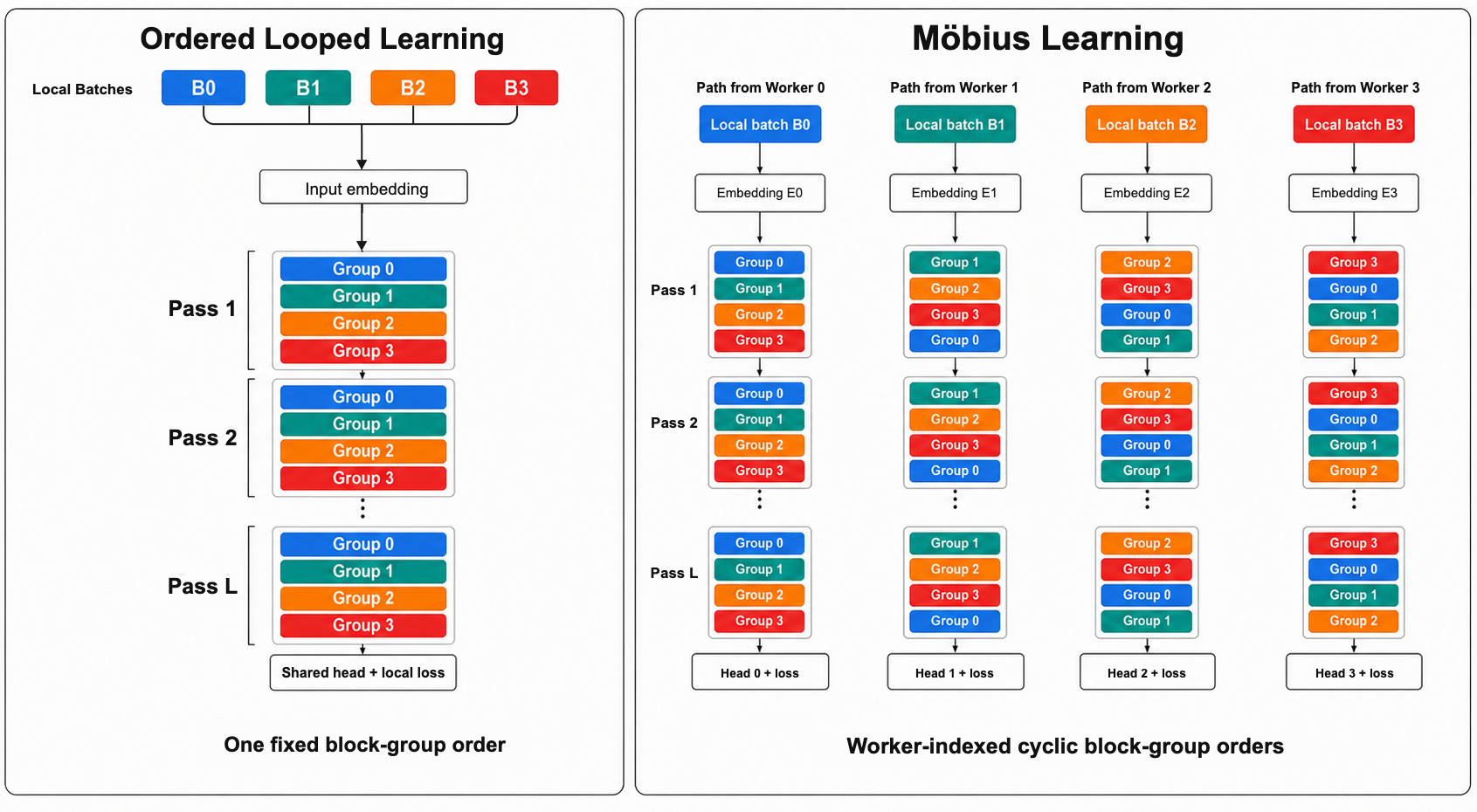}
    \caption{\textbf{Ordered-loop training and \method{} for $P=4$.} \textbf{Left:} every worker-local batch follows the fixed order $F_0,F_1,F_2,F_3$ in each of $L$ complete passes, and the resulting losses are averaged across workers. \textbf{Right:} the column indexed by originating worker $s$ depicts that batch's logical path, which begins at block group $F_s$ and follows the worker-indexed cyclic order; the columns do not represent separate physical copies of the block groups. Along every path, the input embedding is applied once; after $K=LP$ block-group applications, the corresponding prediction map produces vocabulary logits from which cross-entropy is computed.}
    \label{fig:loop-vs-mobius-order}
\end{figure}
\FloatBarrier

\subsection{Source-Relative Depth and Depth-Role Superposition}
\label{subsec:source-relative-depth}

The cyclic shifts induced by different originating workers place the same block group at different positions within a pass. For a batch originating at worker $s$, let $d_{s,i}$ denote the \emph{source-relative depth} of $F_i$: the number of block groups applied before $F_i$ within a pass. We use $\mathcal{T}_{s,L}(i)$ for the positions at which $F_i$ appears in the resulting sequence of $LP$ block-group applications, numbering the first position as zero. Since $F_i$ is first reached when $\langle s+t\rangle_P=i$ and recurs every $P$ applications thereafter, these quantities are
\begin{equation}
    d_{s,i}
    =
    \langle i-s\rangle_P,
    \qquad
    \mathcal{T}_{s,L}(i)
    =
    \left\{
        d_{s,i}+\ell P
        \;:\;
        \ell=0,\ldots,L-1
    \right\}.
    \label{eq:source-relative-depth-set}
\end{equation}
For example, with $P=4$, a batch originating at worker $s=2$ follows $F_2,F_3,F_0,F_1$ within each pass. Hence, $d_{2,0}=2$, and for $L=3$, $\mathcal{T}_{2,3}(0)=\{2,6,10\}$.
For any fixed $i$, the map $s\mapsto d_{s,i}$ is a bijection on $\mathbb{Z}_P$: for any $d\in\mathbb{Z}_P$, its unique inverse is $s=\langle i-d\rangle_P$. Consequently, as the originating-worker index ranges over $\mathbb{Z}_P$, every fixed block group $F_i$ occupies each of the $P$ within-pass source-relative depths exactly once. For a fixed originating worker, that role repeats across the $L$ passes at the positions in $\mathcal{T}_{s,L}(i)$.

This behavior differs from ordered looping, where every worker-local batch encounters $F_i$ at positions $\{i+\ell P:\ell=0,\ldots,L-1\}$. Under cyclic depth folding, batches originating at different workers place the same $F_i$ at every within-pass source-relative depth. The parameters of $F_i$ are therefore trained across all $P$ depth roles, a phenomenon we call \emph{depth-role superposition}.

\subsection{Training and Evaluation Objectives}
\label{subsec:training-evaluation-objectives}

Let worker $s$ draw an input--target pair $(x_s,z_s)$ from its local data distribution $\mathcal{D}_s$. Let $\operatorname{CE}:\mathcal{G}\times\mathcal{Z}\to\mathbb{R}$ denote cross-entropy. The prediction maps $R$ and $R_s$ produce vocabulary logits; $\operatorname{CE}$ converts those logits and the target sequence into a scalar loss.

For ordered-loop training, the input embedding, prediction map, and ordered block-group maps are shared across worker-local data distributions. Its objective is
\begin{equation}
    \min_{E,R,\{F_i\}_{i=0}^{P-1}}
    \frac{1}{P}
    \sum_{s=0}^{P-1}
    \mathbb{E}_{(x_s,z_s)\sim\mathcal{D}_s}
    \left[
        \operatorname{CE}\!\left(
            R\!\left(\Psi^L(E(x_s))\right),
            z_s
        \right)
    \right].
    \label{eq:ordered-training-objective}
\end{equation}

For \method{}, block-group maps remain assigned to their workers, while $E_s$ and $R_s$ denote the maps invoked by each originating path. Let $\Theta$ collect the trainable parameters of the resulting model. Its objective at loop depth $L$ is
\begin{equation}
    \min_{\Theta}
    \frac{1}{P}
    \sum_{s=0}^{P-1}
    \mathbb{E}_{(x_s,z_s)\sim\mathcal{D}_s}
    \left[
        \operatorname{CE}\!\left(
            R_s\!\left(\Phi_{s,LP}(E_s(x_s))\right),
            z_s
        \right)
    \right].
    \label{eq:mobius-training-objective}
\end{equation}
Each originating worker $s$ backpropagates a path-specific loss on a distinct local minibatch. With equal worker participation, \cref{eq:mobius-training-objective} is the resulting population objective; training does not require explicit averaging of these losses. Each block group $F_i$ participates in every term at source-relative depth $d_{s,i}$. Forward evaluation follows \cref{eq:mobius-state-recursion}, and backpropagation traverses the unrolled dependencies in reverse.

Raw input--target pairs remain at their originating workers; hidden states and their gradients cross worker boundaries. Further implementation details appear in \acref{app:evidence-notes}.

\Needspace{12\baselineskip}
For evaluation, let $\mathcal{V}_s$ denote the held-out validation distribution at worker $s$, with all trained parameters held fixed. The ordered-loop criterion is the averaged loss in \cref{eq:ordered-training-objective} with $\mathcal{D}_s$ replaced by $\mathcal{V}_s$. The corresponding \method{} criterion is
\begin{equation}
    \mathcal{L}_{\mathrm{val}}^{\text{M{\"o}bius}}(L)
    =
    \frac{1}{P}
    \sum_{s=0}^{P-1}
    \mathbb{E}_{(x_s,z_s)\sim\mathcal{V}_s}
    \left[
        \operatorname{CE}\!\left(
            R_s\!\left(\Phi_{s,LP}(E_s(x_s))\right),
            z_s
        \right)
    \right].
    \label{eq:mobius-validation-loss}
\end{equation}
Evaluation performs explicit metric aggregation: after each of the $P$ originating-worker paths completes $L$ passes, \cref{eq:mobius-validation-loss} averages their scalar losses to produce the reported validation loss. This aggregation assigns equal weight to all $P$ cyclic starting points and reports performance across the complete set of block orders induced by \method{}, rather than from a selected starting point.

\section{Main Results}
\label{sec:main-results}
\FloatBarrier

\subsection{Comparison Across Loop Depths}
\label{subsec:comparison-loop-depths}

We compare cyclic depth folding with ordered-loop training across loop depths $L$. All runs use four workers, the modded GPT-2 small (124M) architecture~\citep{radford2019language}, and FineWeb~\citep{penedo2024finewebdatasetsdecantingweb}. \method{} divides the Transformer block stack into four worker-resident groups. Following the Chinchilla token-to-parameter allocation for this model scale~\citep{hoffmann2022trainingcomputeoptimal}, we use a 2.5B-token budget; the Muon configuration, learning-rate schedule, and global batch are fixed across runs.

Ordered-loop training repeats a fixed sequence of $P=4$ block groups for $L$ passes; \method{} uses worker-indexed cyclic orders of the same groups. Each table row holds $L$ fixed: both constructions use the same token budget, perform $K=4L$ block-group applications along each input path, and apply the input embedding and prediction map once. Since $K$ increases with $L$, the comparison is matched between methods within each row, not across loop depths.

Both methods are evaluated on the same FineWeb validation data using an equally weighted mean over four losses, one per originating worker. For ordered-loop training, each loss is measured after the final Transformer block-sequence pass; for \method{}, \cref{eq:mobius-validation-loss} averages the path-specific losses after each path completes $K$ block-group applications. \Cref{fig:gb512-validation,tab:gb512-results} show that \method{} reaches lower validation loss at $L=6,10,15$.

\begin{table}[!ht]
\caption{FineWeb validation loss after 2.5B tokens for four-worker modded GPT-2 small (124M) runs. Each row fixes $L$ and $K=4L$; $\Delta$ is \method{} loss minus ordered-loop loss, computed from the unrounded values, so negative values favor \method{}.}
\label{tab:gb512-results}
\centering
\small
\setlength{\tabcolsep}{9pt}
\renewcommand{\arraystretch}{1.18}
\arrayrulecolor{TableRuleGray}
\begin{tabular}{rrrrr}
\toprule
\rowcolor{TableHeaderPurple}
\bfseries $L$ & \bfseries Ordered-loop loss & \bfseries $K$ & \bfseries \method{} loss & \bfseries $\Delta$ \\
\midrule
\cellcolor{TableStructuralPurple}1 & 3.3042 & \cellcolor{TableStructuralPurple}4 & 3.4739 & $+0.1697$ \\
\cellcolor{TableStructuralPurple}2 & 3.2890 & \cellcolor{TableStructuralPurple}8 & 3.3448 & $+0.0557$ \\
\cellcolor{TableStructuralPurple}6 & 3.2564 & \cellcolor{TableStructuralPurple}24 & 3.2480 & $-0.0084$ \\
\cellcolor{TableStructuralPurple}10 & 3.2456 & \cellcolor{TableStructuralPurple}40 & 3.2397 & $-0.0059$ \\
\cellcolor{TableStructuralPurple}15 & 3.2420 & \cellcolor{TableStructuralPurple}60 & \textbf{3.2337} & $-0.0083$ \\
\bottomrule
\end{tabular}
\end{table}
\FloatBarrier

Strikingly, at $L=6,10,15$, the same block parameters serve multiple source-relative depth roles while \method{} attains lower validation loss than ordered-loop training under matched token budgets and block-group application counts.

\FloatBarrier

\section{Applications and Implications}
\label{sec:applications}

\textbf{Representation hierarchy.}
Layer-wise studies show that Transformer representations, attention behavior, and feed-forward functions vary across depth~\citep{tenney2019bertrediscoversclassicalnlp,voita2019bottomupevolutionrepresentationstransformer,clark2019doesbertlookat,geva2021transformerfeedforwardlayerskeyvalue}. The result from \method{} is consistent with these findings while testing a distinct architectural constraint: whether each block group must retain one fixed source-relative depth. Under cyclic depth folding, the same block parameters are trained on representations encountered at shallow and deep source-relative positions, producing depth-role superposition. At loop depths $L=6,10,15$, this construction achieves lower validation loss than ordered-loop training under the same 2.5B-token budget and loop depth. This result raises a mechanistic question: how can one shared block group support both depth roles, and to what extent can their associated representational functions be mixed within the same parameters? We hypothesize that conventional ordered training may leave some reusable capacity untapped by confining each parameter group to one source-relative depth; direct analyses of learned representations and parameter use are needed to test this capacity-utilization account.

\textbf{Worker-local model ownership.}
\method{} combines local data ownership with worker-local ownership of model capacity. Federated learning commonly keeps data local while giving each participant a full model replica~\citep{mcmahan2017communicationefficientlearning}. Pipeline training partitions model capacity across workers, with data sent through a designated input stage. Split learning keeps raw examples local while exchanging activations, usually through fixed split points or stage roles~\citep{vepakomma2018splitlearninghealthdistributed,thapa2022splitfedfederatedlearningmeets}. In \method{}, each worker forms batches from its own data and stores one block group. The resulting hidden states circulate through the block groups in cyclic order, allowing the workers to realize a deeper model collectively. This design opens a concrete path to collaborative training in which each worker trains from local data and contributes a locally held share of model capacity.

\section{Next Version and Future Work}
\label{sec:next-version}

The next version will broaden the current four-worker scaling study. Future experiments will vary the worker count $P$ and loop depth $L$, with $K=LP$ block-group applications for $L$ complete passes, while moving to larger models and longer training-token budgets. This study will characterize when cyclic depth folding achieves lower validation loss than ordered-loop training under the same token budget and loop depth, where the effect saturates, and how it changes with model scale.

A complementary analysis will study how depth-role superposition appears in representations. In \method{}, a block group is not fixed as either shallow or deep; it is repeatedly optimized in both roles as data streams encounter it at different positions in the cyclic order. The mechanism question is therefore broader than any single diagnostic: how do the same parameters organize representations when they must serve multiple positions along Transformer depth? This framing points toward a representation-level account of depth-role superposition and why it can improve training at larger loop depths.

\paragraph{Citation suggestions.}
We welcome pointers to directly relevant work on looped or recurrent Transformers, federated and split learning, and distributed model partitioning at \texttt{raiden@zju.edu.cn}.

\bibliography{references}

\appendix

\section{Implementation Notes}
\label{app:evidence-notes}

The reported implementation follows the state recurrence in \cref{eq:mobius-state-recursion} with $P=4$ workers and assigns block group $F_i$ to worker $i$. Forward computation sends hidden states through the block groups in worker-indexed cyclic order; gradients propagate backward through the same dependency graph. The corresponding prediction map is applied after $K$ block-group applications, and validation averages the resulting path-specific losses. Decoder-block parameters remain on their assigned workers and are not synchronized across block groups; raw data remain at the originating workers while hidden states and gradients are communicated. The reported results use the Version~1 implementation described here. Additional implementation refinements are under evaluation and are excluded from the present claims.

\clearpage
\section{Additional Related Work}
\label{app:additional-related-work}

\subsection{Layer-Order Adaptation and Depth Sharing}
\label{app:layer-order-related-work}

LayerShuffle fine-tunes fixed-order pretrained Vision Transformers under randomly permuted block orders to improve robustness to execution-order perturbations and layer removal at inference~\citep{freiberger2024layershuffleenhancingrobustnessvision}. Its results show that block outputs and predictive contributions can adapt to execution position, although this robustness is accompanied by lower accuracy under the original fixed order. The relation to \method{} can be separated along three dimensions:

\begin{description}[
    style=unboxed,
    leftmargin=0pt,
    labelindent=0pt,
    labelsep=0.5em,
    itemsep=0.35em,
    topsep=0.35em,
    font=\normalfont
]
    \item[\textbf{Training stage.}]
    LayerShuffle begins with blocks already optimized under one fixed order and then adapts them through fine-tuning. \method{} instead imposes cyclic block orders throughout language-model pretraining, without a preceding fixed-order stage. It therefore addresses the more demanding joint problem of learning the language-modeling task while training the same block parameters in shallow and deep roles.

    \item[\textbf{Effective depth.}]
    LayerShuffle applies each distinct block once within a randomly permuted forward pass. \method{} repeatedly traverses the complete block sequence, reaching $K=LP$ block-group applications along each input path and evaluating the construction across a broad range of loop depths.

    \item[\textbf{Order structure and coverage.}]
    LayerShuffle samples arbitrary block permutations independently across minibatches. \method{} uses the $P$ deterministic cyclic shifts selected by the originating-worker index. Across these paths, every block group occupies every within-pass source-relative depth exactly once, yielding balanced supervision across depth roles.
\end{description}

These differences lead to a distinct empirical question. LayerShuffle asks whether a pretrained model can remain functional under arbitrary order perturbations, whereas \method{} asks whether balanced joint pretraining across depth roles can improve fixed-order looped training under matched token budgets and loop depths. The resulting evidence goes beyond order robustness: at the larger tested loop depths, \method{} achieves lower validation loss than the matched ordered-loop baseline.

\end{document}